\documentclass{article}

\usepackage{PRIMEarxiv}

\usepackage[utf8]{inputenc} 
\usepackage[T1]{fontenc}    
\usepackage{hyperref}       
\usepackage{url}            
\usepackage{booktabs}       
\usepackage{amsfonts}       
\usepackage{nicefrac}       
\usepackage{microtype}      
\usepackage{lipsum}
\usepackage{fancyhdr}       
\usepackage{graphicx}       
\usepackage{amsmath}
\usepackage{amssymb}
\usepackage{booktabs}
\usepackage{todonotes}
\usepackage{chngcntr}
\usepackage[numbers,sort&compress]{natbib}
\usepackage{xcolor}
\hypersetup{
    colorlinks,
    linkcolor={red!50!black},
    citecolor={blue!50!black},
    urlcolor={blue!80!black}
}

\begin{document}

\title{\textbf{Developing Linguistic Patterns to Mitigate Inherent Human Bias in Offensive Language Detection}}

\author{
  Toygar Tanyel\,$^{\xi, \varphi}$\;\;\: Besher Alkurdi\,$^{\xi}$\;\;\: Serkan Ayvaz\,$^{\xi, \psi, }$\thanks{Corresponding Author}
\\\\
  $^{\xi}$Department of Computer Engineering, Yildiz Technical University\\
  $^{\varphi}$Department of Biomedical Engineering, Istanbul Technical University\\
  $^{\psi}$Centre for Industrial Software, Maersk Mc-kinney Moeller Institute, University of Southern Denmark\\
\texttt{$^*$seay@mmmi.sdu.dk}
}

\maketitle
\thispagestyle{empty}

\begin{abstract}
With the proliferation of social media, there has been a sharp increase in offensive content, particularly targeting vulnerable groups, exacerbating social problems such as hatred, racism, and sexism. Detecting offensive language use is crucial to prevent offensive language from being widely shared on social media. However, the accurate detection of irony, implication, and various forms of hate speech on social media remains a challenge. Natural language-based deep learning models require extensive training with large, comprehensive, and labeled datasets. Unfortunately, manually creating such datasets is both costly and error-prone. Additionally, the presence of human-bias in offensive language datasets is a major concern for deep learning models. In this paper, we propose a linguistic data augmentation approach to reduce bias in labeling processes, which aims to mitigate the influence of human bias by leveraging the power of machines to improve the accuracy and fairness of labeling processes. This approach has the potential to improve offensive language classification tasks across multiple languages and reduce the prevalence of offensive content on social media.
\end{abstract}


\section{Introduction}
\label{Int}

The prevalence of offensive content on the internet has become an even bigger problem with the widespread use of social media. Content containing offensive expressions targeting vulnerable groups takes on the nature of exacerbating social problems such as hatred, racism, and sexism. Therefore, the detection of offensive language use on social media is an important issue to prevent widespread sharing of offensive content. Research studies on the detection of offensive expression have increased rapidly in recent years \cite{mozafari2022cross,ranasinghe2021multilingual,el2022multilingual,ANAND2023203,plaza2022integrating}. That said, automatically detecting irony, implication, and various forms of hate speech on social media is still an significant challenge.

To accurately detect offensive language use, natural language-based deep learning models require extensive training with large, comprehensive, and labeled datasets. Unfortunately, manually creating such datasets is both costly and error-prone. Furthermore, the use of sarcasm and slang on social media presents additional challenges for accurate labeling. Contextual information and representative training data are essential for models to effectively classify offensive language in sentences.

Recently, a number of datasets have been created and made available to the public for use in offensive language classification tasks in multiple languages \cite{zampieri2020semeval}. Despite significant manual labeling efforts, these datasets are plagued by a variety of issues, including race-based bias, hate speech with profanity, and the use of slang and creative toxic phrases. Additionally, these datasets are often imbalanced, with a disproportionate number of non-offensive samples, making them susceptible to overfitting on the majority non-offensive class.

The presence of human-bias in offensive language datasets is also a critical issue for deep learning models \cite{bender2021dangers,ray2023chatgpt}. Social media data are inherently biased against some vulnerable groups. As deep learning models learn from these datasets, they absorb this bias. Although it may seem theoretically possible to create "value-neutral" language models using the training data, in practice, such models are far from neutral \cite{borji2023categorical}. This "bias" problem can be broadly categorized into three main concepts: prejudice, stereotypes, and discrimination. However, the important question is why natural language models always contain bias that needs to be examined from the perspectives of psychology and sociology. In the subsequent section, we try to address this question further.

Given that human-based bias may not be entirely removable, the research objective is to explore methods of reducing bias in offensive language datasets and models. To address this issue, we proposed a linguistic data augmentation approach to reduce bias in their labeling processes \cite{tanyel2022linguistic}. Moreover, we extend it to multiple languages to validate the generalizability of the approach. This approach aims to mitigate the influence of human bias by leveraging the power of machines to improve the accuracy and fairness of labeling processes.



\section{Understanding unconscious bias in language models: Prejudice, Stereotypes, and Discrimination}

The process of forming our identities is influenced by the social groups we belong to \cite{doi:10.1177/053901847401300204}. Social identity theory posits that individuals who identify with an in-group may seek to enhance their self-image by exhibiting negative attitudes towards out-groups, such as prejudice, stereotypes, and discrimination. Although these forms of bias are interconnected, they can be distinguished from one another. Further categorization of these biases in language models is available in the study by Weidinger et al. \cite{https://doi.org/10.48550/arxiv.2112.04359}. It is crucial to consider the human biases of prejudice, stereotypes, and discrimination before exploring the biases of language models and datasets.

Humans express social norms and categories through languages, and therefore, these norms and categories must be represented accurately in language models for better detection of the patterns. However, technical artifacts such as language models cannot be considered value-neutral, as they reflect and perpetuate the values and norms present in the training data. Thus, it is important to address the risks of social and ethical harm arising from different categories of language models before considering their use in various applications \cite{https://doi.org/10.48550/arxiv.2112.04359}.

\paragraph{Prejudice.} also known as ``advance judgement,'' is an unfounded conclusion that does not hold true in our daily discourse. Prejudices are frequently found in social media portrayals and exist in many spheres of social life. While prejudice is often simplified in daily life, it is a complex, multi-dimensional phenomenon influenced by various factors, such as evolutionary, ontogenetic, sociopolitical, economic, and historical fields \cite{nesne130323}. Due to the biopsychosocial characteristics of human beings, prejudices can appear in many situations.

In language, it is apparent that people can be influenced by their prejudices without realizing it, even in basic criticism cases on social media against random and unknown individuals. What people read in written language can affect them psychologically, putting them in a certain emotional state that triggers unconscious biases.

This is especially consequential in the context of language models as these models are trained on large datasets that often contain human biases and prejudices. These biases can be perpetuated and amplified through the language generated by the models, leading to further harm and discrimination. Therefore, it is vital to address the biases and ethical concerns associated with language models and to develop strategies to mitigate their negative effects. Understanding the complex nature of prejudice and its impact on language use is an important step towards promoting fairness and equality in social interactions, both online and offline.

\paragraph{Stereotyping.} is a common occurrence in social psychology and involves attributing specific characteristics or behaviors to certain groups based on factors like age, occupation, skin color, gender, and other factors. Stereotypes are essentially cognitive ``shortcuts'' that our brains use to make decisions more quickly, which is why they are often accepted without question. However, stereotypes can also be harmful, leading to biased attitudes and behaviors towards certain groups. For instance, people may have preconceived notions about the behavior or abilities of individuals based on their gender or ethnicity, which can lead to discrimination.

In addition to the social and psychological factors that contribute to stereotyping, linguistic factors can also play a role \cite{throughlanguage}. For example, some languages are gender-neutral (e.g., Finnish and Hungarian), while others have grammatical gender that can reinforce gender stereotypes (e.g., German, Spanish and French). Research has shown that the gender elements in a person's native language can affect their cognitive biases and thoughts \cite{effectGender, eGender1, phillips2003can, sera1994grammatical, sera2002language}. Furthermore, although gender-neutral languages do not emphasize genders, they also may contain unconscious bias that are embedded in the words. For instance, the study \cite{beyza2021turk} examined how gender distinctions are encoded in Turkish, a language that is also grammatically gender-neutral. The results uncovered by linguistics and psychology demonstrated the underlying reasons for the bias we encounter in language models.


\paragraph{Discrimination.} is a long-standing problem in human societies and can take many forms. One of the more insidious ways in which discrimination can manifest is through natural language models, which are computational models that use statistical algorithms to understand and generate human language. These models have become ubiquitous in our daily lives, powering everything from virtual assistants to search engines and chatbots. However, as with any human-created system, natural language models are susceptible to biases and discrimination that can be embedded in the data they are trained on \cite{NEURIPS2021_1531beb7}.

The problem of discrimination in language models arises because these models are developed based on huge amounts of text data retrieved from the internet, which contains a plethora of societal biases and prejudices. For instance, language models might be trained on text data that contains racist or sexist language, which can lead to the model associating certain words or phrases with negative stereotypes. As a result, when the model is used to generate text or respond to queries, it may produce outputs that perpetuate harmful stereotypes and discriminatory practices.

Furthermore, language models can also be discriminatory in their output if the training data is not diverse enough to represent different demographics or if the model is biased towards a particular dialect or language variety. This can result in language models that are more accurate for certain groups than for others, which can perpetuate existing power imbalances and inequalities.

The consequences of discriminatory language models can be significant, ranging from perpetuating harmful stereotypes to reinforcing systemic discrimination against marginalized groups \cite{nadeem-etal-2021-stereoset}. As natural language models continue to become more widespread and integrated into our daily lives, it is necessary to address the problem of discrimination in these models and work towards creating more equitable and fair computational systems. This paper explores the causes and consequences of discrimination in language models and propose potential solutions to mitigate the problem.

\section{Related Work}
The proliferation of toxic online language use has become a prominent problem, mainly because of the increasing interactions between online platform users from various social backgrounds. Consequently, the offensive language detection task has gained significant attention lately \cite{ranasinghe2021multilingual,mozafari2022cross,plaza2022integrating,ANAND2023203,el2022multilingual}. This section provides a short overview of the related datasets and models utilized for detecting offensive language, with a particular emphasis on the Turkish and English languages.

\subsection{Offensive language detection datasets and models} 
The problem of offensive language detection has been investigated as a text classification task from diverse perspectives \cite{risch-etal-2020-offensive}. In recent years, a set of datasets has been released to facilitate research in offensive language detection in numerous languages. As a part of Offensive Language Detection (OffensEval) task held in the International Workshop on Semantic Evaluation (Sem-Eval), multiple datasets were provided in five different languages \cite{zampieri2020semeval}. OLID \cite{zampierietal2019} and SOLID \cite{rosenthal2020large}, among others, are well-known datasets used in research for training offensive language detection models in English. In the case of Turkish, a few related datasets were offered recently, including Coltekin's dataset \cite{coltekin-2020-corpus} published in OffensEval-2020, which forms the baseline for the evaluations of in the study, and a dataset offered by Mayda et al. \cite{9599042}, a smaller dataset focusing on Turkish hate speech. 

In \cite{ozdemir2020nlp}, the authors explored various methods and proposed an ensemble model combining LSTM, CNN, BiLSTM, and Attention networks with BERTurk as the underlying language model. Their model was evaluated on Coltekin's dataset. Although their research had similarities to our study, the authors did not report recall as part of their model performance assessment. Therefore, we could not directly compare our model performance results.

In another study, Ozberk and Cicekli investigated the performance effects of different pre-processing techniques on BERT models developed using Turkish text data \cite{ozberk2021offensive}. Moreover, they fine-tuned BERTurk, DistilBERTurk, and ConvBERTurk models on OffensEval-2020 Coltekin's dataset. The study's results indicated that the BERT model was the primary factor affecting performance, with minor effects resulting from pre-processing method changes. Additionally, some research has considered data augmentation methods to address the issue of imbalanced labels in offensive language detection datasets, as is the case in the current study \cite{ekinci2022classification},\cite{rupapara-smote}.

Wei et al.\cite{DBLP:journals/corr/abs-2108-03305} conducted an investigation into commonly used methods for synonym replacement for English based on pre-trained word embeddings, random swapping, random deletion and random insertion. The authors also applied transfer learning to improve the performance results and provided metrics to analyze model performance. Other research has explored the application of BERT models for detection of hate speech on Twitter \cite{zampieri2020semeval}, as well as the use of machine translation for data augmentation \cite{liu2022offensive}.

\subsection{Bias in Large Language Models} 

Bender and Friedman \cite{bender-friedman-2018-data} proposed data statements as a design strategy and professional practice for engineers working in natural language processing technologies. This approach aims to address scientific and ethical challenges associated with using data from certain populations to create technologies for other populations. By adopting and extensively using data declarations, the field may begin to address these critical issues. 

Numerous studies have explored gender bias in depth from a range of perspectives, including surveys \cite{sun-etal-2019-mitigating}, analysis \cite{NEURIPS2020_92650b2e}, and mitigation \cite{mitigate-gender}.
Shah et al. \cite{shah-etal-2020-predictive} provided a comprehensive review of the literature on natural language processing (NLP) and presented more comprehensive mathematical definitions of prediction bias to differentiate between two types of bias outcomes: outcome disparities and error disparities. Additionally, the study identified four potential sources of bias, namely model over-amplification, label bias, selection bias, and semantic bias.

In recent literature, political biases have been investigated as a subject of study \cite{LIU2022103654}, \cite{gangula-etal-2019-detecting}. Nadeem et al. \cite{nadeem-etal-2021-stereoset} proposed a large-scale natural English dataset called StereoSet to measure stereotypical biases in four categories: gender, race, profession, and religion. The authors also evaluated the stereotypical bias and language modeling ability of popular and frequently used models, such as BERT,  RoBERTa, XLnet and GPT-2. The study revealed the strong presence of stereotyped biases in these models.
In order to increase the fairness of language models in social biases, Liang et al. \cite{DBLP:journals/corr/abs-2106-13219} meticulously outlined a number of sources of representational biases, and propose new benchmarks and metrics to quantify biases.

On the other hand, the large language models such as ChatGPT have begun to affect a wide spectrum of aspects of modern society. While they have a strong potential to improve productivity in many areas of life \cite{kasneci2023chatgpt,sallam2023utility}, they are known to have language bias and caution should be exercised when using them \cite{borji2023categorical,bender2021dangers,ray2023chatgpt}.

 
\section{Methodology}	\label{sec:Methodology}
  
\subsection{Overview of the Proposed Approach}

Generating a large dataset through manual human annotation for identifying nuanced linguistic elements in offensive language is both challenging and resource-intensive. To address issues of language bias and the imbalance of labels, which are prevalent in sentiment analysis datasets, this study introduces a methodology that leverages linguistic features to isolate offensive content from an unlabeled corpus. These identified texts are then utilized to supplement the imbalanced dataset, ensuring a more balanced representation across different classes. The efficacy of this approach is evaluated on datasets for detecting offensive language in both Turkish and English.

Hate speech manifests in a variety of forms, targeting different groups including races, nations, religions, individuals, and other entities. Building upon our prior work \cite{tanyel2022linguistic}, this study extends the investigation to identify hate speech linguistic patterns within English dataset. Our approach is notably distinct from synthetic interpolation methods such as ADASYN and SMOTE, as it aims to enhance the context's diversity and generalizability in training data. This enhancement is achieved by incorporating new examples into the existing dataset via data mining. To assess the versatility of our method across languages, we have conducted evaluations using publicly available datasets, focusing on the detection of offensive language in both Turkish and English contexts.

Furthermore, the efficacy of our method was tested by analyzing the performance of NLP models trained using the enhanced dataset. We employed Word2Vec and BERT to examine both statistical and contextual embeddings. In addition, we undertook a comparative analysis of classification performance between SVM, a conventional machine learning model, and CNN-BiLSTM, a more advanced deep learning approach. We have made code related to our method publicly accessible in the project repository \href{https://github.com/tanyelai/lingda}{https://github.com/tanyelai/lingda}.

\subsection{Datasets}

Our baseline dataset was the OffensEval 2020 Turkish Twitter dataset \cite{coltekin-2020-corpus}, consisting of 28,000 samples with a significant imbalance: 22,596 non-offensive and 5,404 offensive samples in the training set, and similar imbalances in the test and validation sets. To address this, we integrated additional datasets, including 5,054 samples from Kaggle, of which 2,073 were offensive, and 10,144 samples from Mayda et al. \cite{9599042}, with 2,502 offensive samples. Further, we mined 13,261 offensive tweets, adding 9,911 to the training set through our proposed method. This improved the balance, resulting in a final training corpus of 42,486 samples, with 22,589 offensive and 19,899 non-offensive labels.

For English, we used the Offensive Language Identification Dataset (OLID) dataset \cite{zampierietal2019}. The dataset contains 14,200 English tweets split into 13,240 train examples and 859 test examples. The annotation process involves three distinct levels: identifying offensive language, categorizing the type of offense, and pinpointing the specific target of the offensive language. We only use the offensive language identification task in this dataset. Similarly to the Turkish dataset, the OLID dataset's label distribution is imbalanced where 4,400 offensive and 8,840 non-offensive texts can be found. We applied numerous variations of our technique based on the arrangement of offensive words, pronouns, and entities, similar to the process followed for Turkish. By applying our method to these different combinations, we obtained a comprehensive dataset that enabled us to show the effectiveness of the approach. 


\subsection{Data Pre-processing}


\subsubsection{Data Cleaning} 
In the data preprocessing phase, the datasets were initially cleansed by removing HTML tags, URLs, emojis, and usernames. Subsequently, the text was converted to lowercase to minimize noise and variation caused by identical words in different cases. Punctuation marks were retained due to their significant impact on BERT embedding distributions. We observed that attention models can utilize the information conveyed by punctuation marks interspersed among words. A notable decline in performance was evident when punctuation marks were excluded from the text. Additionally, it was discerned that certain other preprocessing methods could disrupt BERT's ability to understand contextual relationships. It is important to note that the datasets in our study did not exhibit similar distributions. In simpler terms, the final combined dataset originated from distinctly varied class distributions for the same task. To mitigate the effects arising from differences in class distributions, we merged three diverse datasets for Turkish application and further augmented them using the proposed method \cite{tanyel2022linguistic}.

\subsubsection{Text Normalization} For Turkish language processing, Zemberek, an open source framework, was utilized for text normalization \cite{akin2007zemberek} as demonstrated in Figure \ref{Fig:QueryGeneration}. 
Although text normalization helped improve the performance of models using a statistical word embedding approach, it resulted in worse performance for models using the contextual word embedding approach. As discussed before in \cite{POTA2021115119}, we believe that the main reason for the performance degradation is contextual information lost due to intensive preprocessing. 

As for the English OLID dataset we did not use normalization due to the known negative effect on modern text representation techniques explored in the literature. This decision was supported by the findings of our experiments conducted on the Turkish language.

\subsection{Augmenting Data using Linguistic Features}

To address the challenges posed by label-imbalanced datasets in contextual tasks, adopting a linguistic-based mining approach can reduce the reliance on manually labeling arbitrary data. The initial step in this proposed method involves a thorough understanding of the specific problems and requirements. This approach aims to pinpoint appropriate expressions by leveraging linguistic features, which are effective in identifying the target tweets while maintaining as much of the textual context as possible.

Particularly in tasks like offensive language detection, which are highly reliant on context, datasets are often unbalanced, as also observed in our study. Machine learning models have a tendency to overfit, or become too tailored, to the examples in the majority class. Therefore, addressing the imbalanced nature of the dataset is a crucial preliminary step. This ensures that the model does not develop a bias towards the more frequently represented class and can accurately interpret and classify the data.

In the process of augmenting offensive language datasets, maintaining the natural flow and spontaneity of sentences is essential. To achieve this, the offensive data should not be augmented using methods such as interpolation or weakly supervised generative models. Our proposed method facilitates the automatic retrieval of naturally structured sentences by utilizing a specific set of predefined word combinations.

Additionally, to prevent potential overfitting to non-offensive contexts, it's crucial to expand the dataset with a broad spectrum of natural contexts. This can be achieved by exploring a variety of word combinations. In line with this strategy, and considering the nuances of Turkish linguistics, we employed specific word combinations (phrases) that are particularly suited to the Turkish language. These combinations were carefully selected to ensure they align with the linguistic patterns and syntactic structures inherent in Turkish, thereby enabling the generation of contextually relevant and linguistically coherent data for our study.
\[ query = swear + entity\begin{cases}
      \text{suffix}_{singular}(entity) \\
      \text{suffix}_{plural}(entity)
\end{cases}  \]

To support this approach, we manually compiled two separate lists: one containing common swear words, and another listing entities commonly targeted by offensive language, such as persons, races, religions, and adjectives. These lists were then used to generate the aforementioned word combinations, ensuring a comprehensive and targeted approach to augmenting the dataset.

\begin{figure}[h!]
\begin{center}
\includegraphics[width=\linewidth]{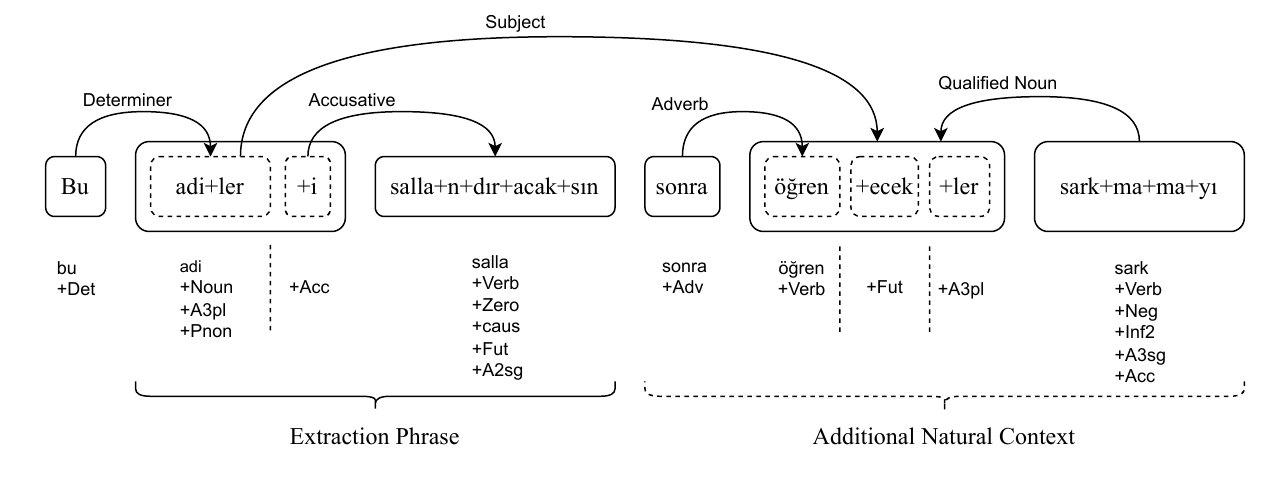}
\caption{Sample query result \cite{tanyel2022linguistic}. Extraction phase highlights the retrieval of the primary context, with the incidental acquisition of additional natural context, obtained without further inference.} \label{Fig:QueryGeneration}
\medskip
\end{center}
\small 
\end{figure}

As shown in Figure \ref{Fig:QueryGeneration}, offensive contexts were extracted from the tweets. Then, the tweets were analyzed morphologically. The sample tweet illustrated in the figure translates to ``We should hang those shitboxes, so that they learn not to rackjack''.

Because of the widespread use of swearwords as a way of implying strong expressions, emotions or sarcasm as discussed in \cite{jonsson2018swedes}, swearwords and entities alone or even in combination are not sufficient for a tweet to be automatically labeled offensive with high confidence. 

\begin{figure}[h!]
\begin{center}
\includegraphics[width=\linewidth]{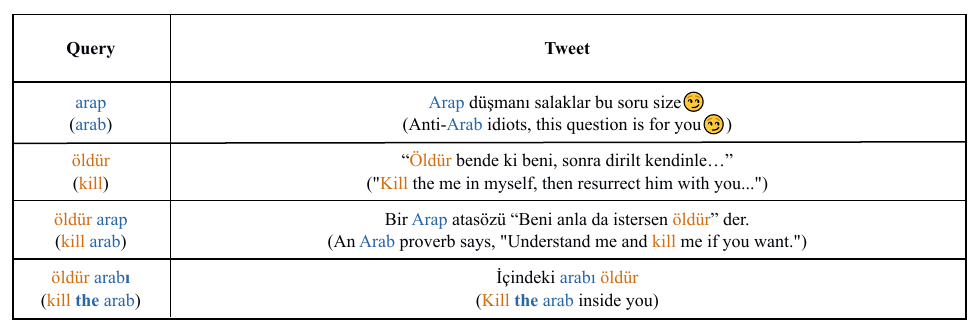}
\caption{This sample illustrates the importance of linguistic features in context retrieval, using the query `öldür arab\textbf{ı}'. The alteration of `Arap' to `Ara\textbf{b}ı' in the presence of a vowel highlights a linguistic phenomenon. 
} \label{Fig:TweetsandQueries}
\medskip
\end{center}
\end{figure}

As shown in the Figure \ref{Fig:TweetsandQueries}, the outcomes of a query examples using `Arap' (Arab) as the entity and `öldür' (kill) as the direct action towards the entity, both of which are potential indicators of hate speech. Analysis of the selected tweets reveals that combinations of hate speech terms with entities do not invariably result in offensive classifications. 
Here, derogatory or hateful expressions are often appended to sentences, intensifying the expression without conveying a direct, explicit meaning \cite{wang2014cursing}. In this study, we employed the aforementioned methodology to generate a comprehensive list of over 3,000 queries. These queries were then utilized to extract tweets that were automatically categorized as offensive

\begin{figure}[tb!]
\begin{center}
\includegraphics[width=\linewidth]{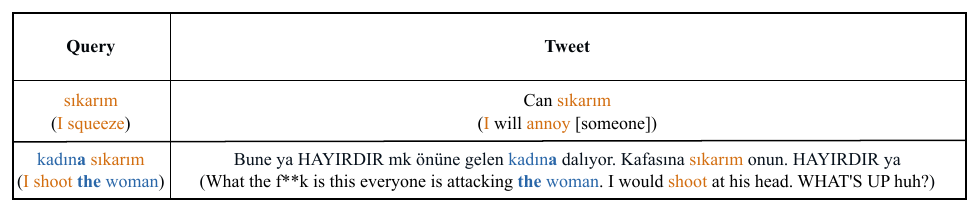}
\caption{A sample of tweets containing homophones} \label{Fig:Homonyms}
\end{center}
\end{figure}

Our evaluations also showed that the proposed method is applicable when combining homophones with entities that can be used in a non-offensive context. As shown in the example in the Figure \ref{Fig:Homonyms}, using these words alone as queries results in tweets that cannot be confidently labeled as offensive.


Additionally, swearwords or potential offensive words may not always be directed at the entity in queries. We also observed that tweets containing entities involving directed objects along with potentially offensive words tended to be offensive in almost all cases in the evaluations. To evaluate this observation, a sample of one hundred tweets were queried from Twitter automatically using the proposed method as described previously, and two annotators were asked to label the tweets as offensive or non-offensive. The results demonstrated that only 2\% of tweets were labeled as non-offensive by both annotators; This shows that our method is effective and scalable.


\subsection{Assessing the Cross-Linguistic Applicability of Our Method}
\begin{figure}[htb!]
\begin{center}
\includegraphics[width=\linewidth]{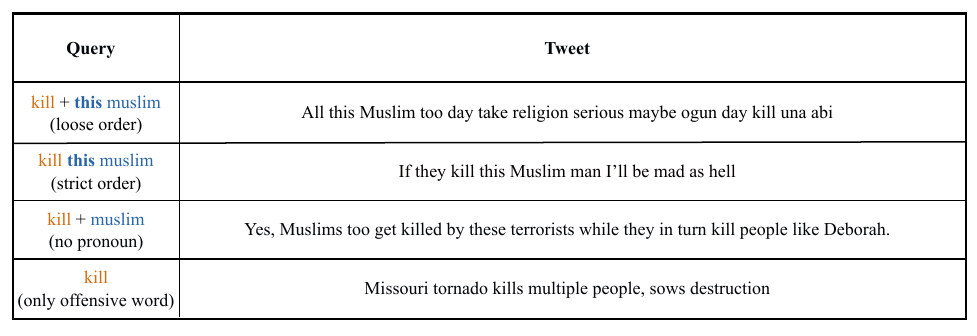}
\caption{Examples of tweets queried using different mining techniques in English.} \label{Fig:tweets_en}
\end{center}
\end{figure}

To investigate the applicability of our approach to other languages, we applied the proposed method to English language tweets using a list of commonly used offensive words and entities. We used the OLID dataset \cite{zampierietal2019} for the evaluations by adding offensive tweets until the number of samples in both classes was equal. We explored various mining techniques, characterized by the arrangement of $OW$, $P$, and $E$, denoting offensive word, pronoun, and entity, respectively. The $+$ operator allows for changes in the order of the elements. We analyzed the loose order method $(OW + P\ E)$, where the order is flexible, and the strict order method $(OW\ P\ E)$, where the order remained fixed. Additionally, we considered methods that exclude pronouns $(OW + E)$ and involve only the offensive word $(OW)$. Figure \ref{Fig:tweets_en} demonstrates examples of tweets queried using the different methods in English.

\subsection{Models}
In this section, we offer a concise overview of the technical aspects of the word embedding technique, the language model, and the training models selected for assessing the datasets. We explain the rationale behind employing both statistical word embeddings and a contextual language model. Furthermore, we delve into the intuition underlying certain distinctions between traditional machine learning and deep learning models in the context of natural language tasks.

\subsubsection{Word2Vec} Word2Vec is an NLP method that is used for building vector based word embeddings from a large text corpus. In the study, we trained Word2Vec model on our Turkish training dataset using Gensim library v4.1.2. A window size of 7 and a vector size of 300 and 16 epochs were selected for model training. We trained another Word2Vec (word2vec\textsubscript{large}) model using the same parameters on tweets collected by querying Twitter using a Turkish word list\footnote{https://github.com/mertemin/turkish-word-list} based on VikiSözlük\footnote{https://tr.wiktionary.org}.

\subsubsection{BERT} BERT (Bidirectional Encoder Representations from Transformers) is the NLP technique that rely on self-attention mechanism. We utilized default BERT (bert-base-uncased) for English, and BERTurk \cite{stefan_schweter_2020_3770924}, a Turkish BERT model with 128k uncased vocabulary. BERT is employed for the conversion of word tokens within textual data into contextual word embeddings. Its superior performance in prediction accuracy compared to other embedding models such as Word2Vec can be attributed to a fundamental difference. Unlike Word2Vec, which assigns a static representation to each word, irrespective of its contextual variations in different contexts, BERT generates dynamic word embeddings that adapt to the surrounding words. This adaptability results in more precise and natural predictions.

\subsubsection{SVM} Support Vector Machine (SVM) is a well-known machine learning algorithm that is commonly used for classification tasks. SVMs are capable of handling high-dimensional data and have been found to perform well in text classification tasks. In our study, we utilized the LinearSVC implementation of SVM from the scikit-learn library \cite{scikit-learn}, training it to minimize the loss function below:

\begin{equation}
\text{Loss} = \frac{1}{2} \sum_{i=1}^{n}{\max(0, 1-y_i \cdot \widehat{y}_i)^2} + \lambda \sum_{i=1}^{n}{w_i^2}
\end{equation}

where ${w_i}$ denotes normal vector to the separation hyperplane and ${y_i} \in {\{1, -1\}}$ represents the actual class for given input point while  ${\widehat{y}_i}$ represents the predicted class for the point. Default $\lambda$ value was used as the regularization parameter.

\subsubsection{CNN} In the realm of text processing, 1D CNNs are recognized for their proficiency in extracting a broad range of features. A notable characteristic of CNN layers is their lack of memory retention, enabling them to operate at high speeds without storing information for future use. Among various embeddings, BERT embeddings are particularly well-suited for CNNs due to their favorable feature distribution. BERT surpasses traditional statistical models by providing a richer context for text analysis. In our study, the integration of CNN layers with BERT's contextually-rich embeddings is an effective strategy. This combination allows for the extraction of nuanced meanings from text, leveraging the strengths of both CNNs and BERT's advanced contextual capabilities.
 
\subsubsection{BiLSTM} 
The BiLSTM model is a sequential processing architecture comprising two Long Short-Term Memory (LSTM) \cite{lstm} units, with one designed to handle input in the forward direction and the other in the reverse direction. This bidirectional nature of BiLSTM enhances the algorithm's contextual awareness, thereby augmenting the volume of information accessible to the neural network.

\section{Results}

\subsection{Evaluation Metrics}
To assess the performance of offensive language detection models, we conducted several evaluations in English and Turkish datasets. 
Since the offensive language datasets are often imbalanced, using Macro F1 is typically the default metric choice. Considering the objective of offensive language detection task, it was more critical for us to focus on increasing the Recall results. Recall measures the ratio of finding relevant data within a dataset. In our case, this refers to the model's ability to identify all instances of offensive language in the target dataset. In other words, the recall score only shows the proportion of ``offensive'' samples that were correctly labeled. Recall$\mathrm{_{avg}}$ measures the macro average recall score of both ``not-offensive'' and ``offensive'' labels. 
As a result, in the offensive language detection task, correctly detecting offensive texts is more important than non-offensive comments because failure to detect toxic content is more harmful.
The F1$\mathrm{_{avg}}$ score represents the macro average results of F1 scores.

\subsection{Training and Performance Evaluations}

To understand how different model contexts affect performance, we evaluated statistical and attention-based approaches using multiple models, including Word2Vec-SVM, Word2Vec$\mathrm{_{large}}$-SVM, BERT-SVM, and BERT-CNN-BiLSTM. These experiments were executed using Python 3.7.13 and TensorFlow 2.8.0, leveraging the GPU and TPU capabilities of Colab Notebook.

\subsubsection{Turkish Model Pipeline using BERTurk}
In our evaluation process, we segmented the dataset into three parts: training, validation, and test sets. The validation set was used to determine the point at which the model begins to learn and when it reaches its saturation point. For the Turkish dataset, the training set consisted of 42,398 samples, the validation set consisted of 1,756 samples, and the test set consisted of 8,851 samples. We used the \textit{"dbmdz/bert-base-turkish-128k-uncased"} model to generate BERT embeddings with a max length parameter of 200 dimensions. The CNN module used 32 dimensions and a filter size of 3, while the BiLSTM had 100 layers and a 0.2 dropout rate in both directions. The training process spanned 3 epochs for the baseline dataset and extended to 18 epochs for our augmented dataset. Parameters included a learning rate of 1e-5, the utilization of an Adam optimizer, and a batch size of 128.  Notably, we observed that variations in batch sizes and learning rates resulted in overfitting or divergence in the models across both datasets.

\renewcommand{\arraystretch}{1.2}
\begin{table*}[htb!]
\caption{Table showing performance results for different models in Turkish \cite{tanyel2022linguistic}} 
\centering
\resizebox{\textwidth}{!}{%
\begin{tabular}{@{}lcccccccccccc@{}}
   \specialrule{.08em}{.05em}{.05em} 

\textbf{Test A} &
  \multicolumn{6}{c|}{\textbf{Without Text Normalization}} &
  \multicolumn{6}{c}{\textbf{With Text Normalization}} \\    \specialrule{.08em}{.05em}{.05em} 

 &
  \multicolumn{3}{c|}{Coltekin's Dataset} &
  \multicolumn{3}{c|}{Our Dataset} &
  \multicolumn{3}{c|}{Coltekin's Dataset} &
  \multicolumn{3}{c}{Our Dataset} \\ \cmidrule(l){2-13} 
Model &
  Recall &
  Recall\textsubscript{avg} &
  \multicolumn{1}{c|}{F1$\mathrm{_{avg}}$} &
  Recall &
  Recall\textsubscript{avg} &
  \multicolumn{1}{c|}{F1$\mathrm{_{avg}}$} &
  Recall &
  Recall\textsubscript{avg} &
  \multicolumn{1}{c|}{F1$\mathrm{_{avg}}$} &
  Recall &
  Recall\textsubscript{avg} &
  F1$\mathrm{_{avg}}$ \\ \midrule
Word2Vec-SVM &
  20.39 &
  59.80 &
  \multicolumn{1}{c|}{61.72} &
  33.94 &
  64.59 &
  \multicolumn{1}{c|}{67.14} &
  29.05 &
  63.85 &
  \multicolumn{1}{c|}{67.14} &
  40.08 &
  67.62 &
  70.42 \\
Word2Vec$\mathrm{_{large}}$-SVM &
  23.04 &
  60.88 &
  \multicolumn{1}{c|}{63.21} &
  38.97 &
  66.71 &
  \multicolumn{1}{c|}{69.22} &
  27.65 &
  63.06 &
  \multicolumn{1}{c|}{66.09} &
  40.92 &
  67.67 &
  70.21 \\
BERT-SVM &
  44.83 &
  70.99 &
  \multicolumn{1}{c|}{74.74} &
  57.96 &
  75.37 &
  \multicolumn{1}{c|}{76.71} &
  43.58 &
  70.17 &
  \multicolumn{1}{c|}{73.71} &
  56.28 &
  74.80 &
  76.48 \\
BERT-CNN-BiLSTM &
  68.85 &
  81.05 &
  \multicolumn{1}{c|}{\textbf{81.59}} &
  \textbf{86.31} &
  \textbf{83.80} &
  \multicolumn{1}{c|}{77.22} &
  72.07 &
  80.45 &
  \multicolumn{1}{c|}{\textbf{81.59}} &
  72.91 &
  82.03 &
  81.13 \\ \midrule
Coltekin Sub-Task A &
  - &
  76.20 &
  \multicolumn{1}{c|}{77.30} &
  - &
  - &
  \multicolumn{1}{c|}{-} &
  - &
  - &
  \multicolumn{1}{c|}{-} &
  - &
  - &
  - \\ \midrule
\multicolumn{13}{l}{} \\    \specialrule{.08em}{.05em}{.05em} 

\textbf{Test B} &
  \multicolumn{6}{c|}{\textbf{Without Text Normalization}} &
  \multicolumn{6}{c}{\textbf{With Text Normalization}} \\    \specialrule{.08em}{.05em}{.05em} 

 &
  \multicolumn{3}{c|}{Coltekin's Dataset} &
  \multicolumn{3}{c|}{Our Dataset} &
  \multicolumn{3}{c|}{Coltekin's Dataset} &
  \multicolumn{3}{c}{Our Dataset} \\ \cmidrule(l){2-13} 
Model &
  Recall &
  Recall\textsubscript{avg} &
  \multicolumn{1}{c|}{F1$\mathrm{_{avg}}$} &
  Recall &
  Recall\textsubscript{avg} &
  \multicolumn{1}{c|}{F1$\mathrm{_{avg}}$} &
  Recall &
  Recall\textsubscript{avg} &
  \multicolumn{1}{c|}{F1$\mathrm{_{avg}}$} &
  Recall &
  Recall\textsubscript{avg} &
  F1$\mathrm{_{avg}}$ \\ \midrule
Word2Vec-SVM &
  69.81 &
  84.60 &
  \multicolumn{1}{c|}{84.28} &
  81.86 &
  88.66 &
  \multicolumn{1}{c|}{88.62} &
  76.81 &
  87.86 &
  \multicolumn{1}{c|}{87.73} &
  84.55 &
  89.99 &
  89.97 \\
Word2Vec$\mathrm{_{large}}$-SVM &
  75.20 &
  87.11 &
  \multicolumn{1}{c|}{86.95} &
  84.21 &
  89.54 &
  \multicolumn{1}{c|}{89.52} &
  76.40 &
  87.60 &
  \multicolumn{1}{c|}{87.47} &
  84.78 &
  89.85 &
  89.84 \\
BERT-SVM &
  77.40 &
  87.46 &
  \multicolumn{1}{c|}{87.35} &
  87.61 &
  90.25 &
  \multicolumn{1}{c|}{90.25} &
  76.87 &
  87.07 &
  \multicolumn{1}{c|}{86.96} &
  87.13 &
  90.29 &
  90.28 \\
BERT-CNN-BiLSTM &
  91.12 &
  92.22 &
  \multicolumn{1}{c|}{92.23} &
  \textbf{96.47} &
  88.88 &
  \multicolumn{1}{c|}{88.80} &
  92.91 &
  \textbf{92.48} &
  \multicolumn{1}{c|}{\textbf{92.48}} &
  93.18 &
  92.35 &
  92.35 \\ \bottomrule
\end{tabular}%
}
\label{tab:performance}
\end{table*}

Table \ref{tab:performance} demonstrates the performance evaluation results in terms of macro recall of offensive labels, recall$\mathrm{_{avg}}$ and F1 scores. In the table, Test A represents the baseline evaluations using Coltekin's test set. Test B represents evaluation results using our newly created test set, which includes the dataset from Test A and randomly selected samples from the new dataset. Our Dataset columns represent the performance results of models trained using our newly proposed dataset. The impact of text normalization was also investigated as an evaluation parameter. The text normalization columns in the table indicate whether or not text normalization was applied in the evaluation.

\subsubsection{English Model Pipeline using BERT}
For the English Model pipeline, the dataset was similarly divided into training, validation, and test sets, with the validation set serving the same purpose as in the Turkish model. Here, the training set comprised 13,240 samples, while the test set contained 859 samples. We employed the "bert-base-uncased" model for BERT embeddings, maintaining the maximum length parameter at 200 dimensions. The model configuration mirrored that of the Turkish pipeline, with a 32-dimension CNN module, a filter size of 3, and a BiLSTM with 100 layers and a 0.2 dropout rate. The training duration was set at 5 epochs. Consistent with the Turkish model, we used a 1e-5 learning rate, an Adam optimizer, and a batch size of 128. 

Table \ref{tab:olidenglish} demonstrates the performance results of different models in English using OLID dataset. In the table, recall, recall average, and F1 average metrics are presented for Word2Vec-SVM and BERT-CNN-BiLSTM models, with various modifications to the mining method. The loose order showed the highest positive recall score. No data augmentation was applied for baseline.

Similar to our findings from experiments on the Turkish dataset, augmenting retrieved tweets led to improved results based on the positive recall metric, this comes with a trade-off in accuracy. Our investigations showed that the loose order method provided the best performance improvement, consistent with the experiments conducted in Turkish. We were able to retrieve a significant number of offensive tweets using our method, demonstrating its effectiveness in identifying potentially offensive content in English language tweets.


\renewcommand{\arraystretch}{1.2}
\begin{table*}[t]
\caption{Performance results of different models tested on the OLID test split in English} 
\centering
\resizebox{0.85\textwidth}{!}{%
\begin{tabular}{@{}clccccccc@{}}
\toprule
\textbf{}           & \textbf{} & \multicolumn{3}{c}{\textbf{Word2Vec-SVM}}  &  & \multicolumn{3}{c}{\textbf{BERT-CNN-BiLSTM}} \\ \cmidrule(l){3-9} 
\textbf{Mining Method} &
   &
  \textbf{Recall} &
  \textbf{Recall$_\mathrm{avg}$} &
  \textbf{F1$_\mathrm{avg}$} &
   &
  \textbf{Recall} &
  \textbf{Recall$_\mathrm{avg}$} &
  \textbf{F1$_\mathrm{avg}$} \\ \cmidrule(r){1-1} \cmidrule(lr){3-5} \cmidrule(l){7-9} 
Baseline            &           & 44.35 ± 0    & 69.03 ± 0    & \textbf{71.16 ± 0}    &  & 70.5          & \textbf{79.6}          & \textbf{80.6}         \\
Loose Order         &           & \textbf{58.66 ± 0.35} & 69.27 ± 0.43 & 68.61 ± 0.51 &  & \textbf{78.6}          & 78.6          & 75.8         \\
Strict Order        &           & 56.9 ± 1.51  & \textbf{70.12 ± 1.14} & 70.11 ± 1.23 &  & 73.6          & 79.2          & 78.2         \\
No Pronoun          &           & 57.53 ± 0.89 & 69.01 ± 0.45 & 68.52 ± 0.38 &  & 76.4          & 76.8          & 73.8         \\
Only Offensive Word &           & 55.65 ± 0    & 65.24 ± 0.34 & 64.18 ± 0.42 &  & 70            & 79            & 79.4         \\ \bottomrule
\end{tabular}}
\vspace{0.25cm}
\label{tab:olidenglish}
\end{table*}

\subsubsection{Performance Analysis of Models}
Our comparative evaluations showed that while Word2Vec and Word2Vec$\mathrm{_{large}}$ models provided inconsistent results, Word2Vec$\mathrm{_{large}}$ performed better than Word2Vec when text normalization was not used, but attention-based models still outperformed it. SVM is a faster alternative to neural networks, and we evaluated it with two word embeddings and a language model. Word2Vec did not provide enough word connections for SVM to be effective, while BERT offered more contextual information, explaining the difference in performance between Word2Vec-SVM and BERT-SVM on Test A. The BERT-CNN-BiLSTM pipeline outperformed the other models in both test cases, using BERT's extensive feature extraction, CNN's feature discovery, and BiLSTM's ability to understand word connections. Text normalization was beneficial for statistical models like Word2Vec, but had a negative impact on BERT, which can handle misspelled and unknown words and depends heavily on sentence context. Overall, the results demonstrated the advantage of providing more context with the dataset, and the superiority of the BERT-CNN-BiLSTM pipeline.

\subsection{Results of Error Analyses}

\subsubsection{Error Analysis of Turkish Dataset}



\begin{table}[h!]
   \caption{Examples of misclassified samples in the Turkish dataset}
   \label{tab:wrongDevRevised}
   \small
   \centering
   \renewcommand{\arraystretch}{1.5}
   \begin{tabular}{|c | p{5cm} | c | c|}
    \hline
    \textbf{ID} & \textbf{Text} & \textbf{Label} & \textbf{Predicted} \\ \hline
  
   A & ata'ya laf edenlerin hepsi sürülecek, laiklik karşıtlarına yer yok.\newline
   \textit{(anyone who speaks against the father [Atatürk] will be exiled, no place for opponents of secularism.)} 
   & NOT & OFF \\ \hline
   B & ``havaya atılan bir taş düşünebilseydi kendi isteğiyle yere düştüğünü sanırdı'' demiş Spinoza üstad.\newline
  \textit{(``if a stone thrown into the air could think, it would believe that it fell to the ground of its own will,'' said master spinoza.)}
   & NOT & OFF \\ \hline 
   
   C & herkes kendi kültüründe kalmalı, böylece her kültür kendi içinde gelişir ve korunur.\newline
    \textit{(everyone should stick to their own culture, so that each culture can develop and be preserved within its own.)}
    & OFF & NOT \\ \hline
   
   D & o zaman da salak gibi kararsızdım\newline
   \textit{(back then, I was stupidly indecisive.)}
   & OFF & NOT \\ \hline
   
   E & hepsi ceplerini çok güzel dolduruyor vatandaşta birbirine sayıyor işte\newline
   \textit{(they all fill their pockets very well, and the citizens end up cursing each other.)}
   & OFF & NOT \\ \hline 
   
   \end{tabular}
\end{table}

In Table \ref{tab:wrongDevRevised}, we present examples in Turkish, highlighting differences between actual labels and predictions by a BERT-CNN-BiLSTM model. Examples A, D, and E are incorrectly labeled as offensive in the dataset. Samples A and E, while politically assertive, lacks offensive language. Sample D, though critical in tone, is not inherently offensive. 

For Samples B and C, labeled correctly in the dataset but misclassified by the model, we hypothesize the following: Sample B, a philosophical musing, likely confused the model due to the usage of the words "throwing" and "stone". Sample C, a suggestion in culture context, may have been deemed non-offensive by the model due to its innocuous framing, overlooking subtle offensive implications. These examples underscore the challenges in sentiment analysis, especially in capturing the nuances of language and context.

\subsubsection{Error Analysis of The OLID English Dataset}

This analysis focuses on the misclassified examples by the BERT-CNN-BiLSTM pipeline in the OLID dataset, highlighting the model's struggle in differentiating between offensive and non-offensive content.

\begin{table}[h!]
  \caption{Examples of misclassified samples in English dataset}
  \label{tab:wrongEngDev}
  \small
  \centering
  \renewcommand{\arraystretch}{1.5}
  \begin{tabular}{|c | p{5cm} | c | c|}
  \hline
    \textbf{ID} & \textbf{Text} & \textbf{Label} & \textbf{Predicted} \\ \hline
    
    A & His fans are happy. He spreads love and because of him I learned to appreciate and support the LGBTQ. I love him because he never failed to make me happy. I love him because he is Harry Styles and Harry Styles is the best.
    & NOT & OFF \\ \hline
    
    B & Amazon will ship live Christmas trees to your door — but will they stay and put on the damn lights?
    & OFF & NOT \\ \hline
    
    C & \#kalyani is a very stupid teenager. I totally despise her for the atrocities she is doing towards aunty Anupriya. She blames her for her mom’s death. She should stop and listen to Anupriya side of the story before she makes vile accusation.
    & OFF & NOT \\ \hline
    
    D & \#california patriots... especially those in \#ca21 Obama/Soros are attempting to take this house seat to get control of congress. Say no to TJ Cox. He is a shill candidate fronted by the dastardly duo. Please get out and vote David Valadao for house, dist 21!
    & OFF & NOT \\ \hline
    
    E & \#media \#credentials if you are POC and had a problem with please send DM me do not like this tweet
    & NOT & OFF \\ \hline
  \end{tabular}
\end{table}

This table shows a selection of examples from the English OLID test set that were misclassified by the model. In the table, "OFF" denotes offensive, and "NOT" denotes non-offensive content. The chosen examples highlight different types of errors:

\begin{enumerate}
    \item \textbf{Misinterpretation of Positive Sentiment (Example A and E)}: The model incorrectly classified tweets with positive or supportive sentiments as offensive. This indicates a challenge in interpreting context and emotional tone, especially when certain keywords typically associated with offensive language in the training dataset are present.

    \item \textbf{Overlooking Subtle Offensiveness (Example B and C)}: The model failed to identify subtle offensive content, such as sarcastic remarks or indirect insults. This highlights the difficulty in detecting nuanced expressions of offensiveness that rely heavily on contextual understanding and cultural nuances.

    \item \textbf{Political Bias (Example D)}: The model's inability to correctly classify certain politically charged tweets suggests a challenge in handling politically oriented language. This could stem from biases in the training data or the inherent complexity of political discourse.
\end{enumerate}

The misclassification patterns observed in the OLID English dataset underscore the importance of context and the subtlety of language in offensive content detection. It is evident that while the BERT-CNN-BiLSTM pipeline is effective in many cases, it struggles with nuanced expressions, emotional tone, and politically charged language. To improve the model's performance, future work should focus on enhancing its ability to understand context and subtle language variations. Ensuring a balanced representation of different types of content in the training data could help reduce biases and improve overall precision.


\section{Discussion}

Exposure to toxic content is a very common problem on the internet, especially on social media \cite{beyza2021turk}. Accurate detection of toxic content is important for the implementation of preventive measures. Deep natural language processing models often contain inherent biases in the data, as they are developed by collecting large human-generated datasets \cite{abid2021persistent}. These biases are likely to mislead the models developed for use in detecting toxic content on the internet \cite{nadeem-etal-2021-stereoset}. 



Unconscious bias, which includes prejudice, stereotypes, and discrimination, can be perpetuated through language models \cite{NEURIPS2021_1531beb7}. Prejudice is an unfounded conclusion that does not hold true in our daily discourse, and it is a complex, multi-dimensional phenomenon influenced by various factors. Stereotypes are cognitive “shortcuts” that our brains use to make decisions more quickly, and they can be harmful, leading to biased attitudes and behaviors towards certain groups. Discrimination, on the other hand, is a long-standing problem in human societies and can take many forms.

To mitigate the negative effects of language models, researchers can develop methods to identify and remove biased language from training data, or create models that are specifically designed to be fair and unbiased \cite{liu2021mitigating}. For example, language models can be tested for bias and then fine-tuned to reduce or eliminate any biases that are detected. Additionally, diversity and inclusivity can be prioritized when selecting and curating training data. By taking steps to address the biases and ethical concerns associated with language models, we can promote fairness and equality in social interactions, both online and offline \cite{bender-friedman-2018-data}.

The inherent language bias in natural languages manifests itself in imbalanced datasets and leaks into the language models. Addressing latent biases in the data itself is critical for developing unbiased large-language models. Thus, bias should be reduced on the data side first. 

Our approach focuses on reducing the stereotypical bias in the offensive language detection models by augmenting the data using linguistic features. By incorporating more contextual information, the predictions become more generalized and accurate, even for extreme cases. We have demonstrated that simple language features can be used to retrieve contextual text and address label imbalances for tasks such as offensive language detection.

It is also worth noting that our approach can be generalized to multiple languages. To assess the cross-language applicability of the method, we collected multiple data sets in Turkish and English. Evaluations showed that the proposed approach was applicable to both Turkish and English.

We noticed that our approach can be beneficial not only to increase the size of data in cases where data quality is important and the labeling process is expensive, but also to reduce errors in the labeling process. Additionally, we highlighted the challenges associated with selecting a scoring metric for imbalanced datasets. Our method improved the macro average recall score by 7.60\% from the compared baseline. Moreover, the offensive recall score increased significantly from 68.85\% to 86.31\%, an increase of 17.46\% compared to the baseline test set in Turkish. 

Similarly, in English, using the BERT-CNN-BiLSTM pipeline, we observed that recall scores increased by 8.1\% from baseline. However, F1 scores were similar or slightly lower than baseline due to lower precision results. Due to the imbalanced nature of offensive text classification datasets, we observed a decrease in precision scores when offensive content in the dataset was augmented.
That being said, given the trade-off between precision and recall, recall is more important in our case; because capturing offensive content and mitigating bias in text by increasing the representativeness of offensive content is more crucial than correctly predicting non-offensive text.

It should be more discussed that from a methodological perspective, we compared statistical embedding method, Word2Vec and contextual embedding method of BERT using a classification modeling pipeline. Our tests showed that the context-based embedding approach works better than the statistical embedding approach. However, another important finding we obtained in the study was that the text normalization approach does not always provide a positive effect. While text normalization led to better results in the statistical approach, it negatively affected the contextual embedding approach. Text normalization appears to affect the context of the sentences. 
 
 

The augmented datasets have still potential for improvement. The precision-recall trade-off is evident due to the limited focus of the augmentation on offensive texts. To address this, increasing the size of available data could be helpful to achieve better results. Additionally, the produced dataset may still contain racial bias, as it only includes offensive context about races. To overcome this problem, future work could use our approach to augment the data with positive context for a list of entities. This will allow models trained on the data to learn from both positive and negative examples associated with the entity. Thus, it will be avoided to classify sentences based only on biased mentions of entities. 


\section{Conclusion}

Large language models, such as those used in virtual assistants, search engines, and chatbots, are based on vast amounts of text data from the internet. This data is not neutral, and it reflects societal biases and prejudices. Consequently, language models can perpetuate stereotypes, biases, and discriminatory practices, leading to negative impacts on marginalized groups. Therefore, it is essential to develop strategies to address the ethical and social risks associated with language models and mitigate their negative effects. The datasets used in training language models for offensive language detection often present significant issues such as being imbalanced, containing many mislabeled samples and being racially biased. 

In this study, we developed a linguistic data augmentation approach that aims to automatically reduce the impact of human bias in labeling processes by extracting high-level language features. The performance of the data augmentation approach was evaluated on both Turkish and English offensive language detection datasets using Word2Vec-SVM and BERT-CNN-BiLSTM models. Additionally, we evaluated the effects of selecting text normalization, word embedding type, language model and compared the classification performance of SVM, a traditional machine learning method, with the CNN-BiLSTM model, a hybrid deep learning model. The metric focus was shifted from F1 to macro recall of offensive label to better suit real-life usage. Based on the Turkish and English performance evaluation of our approach, we found that this approach can help improve offensive language classification tasks across multiple languages and reduce the prevalence of offensive content on social media.

{\small
\bibliographystyle{unsrt}
\bibliography{refs}
}

\end{document}